\documentclass[conference,a4paper]{IEEEtran}
\IEEEoverridecommandlockouts
\usepackage{cite}
\usepackage{amsmath,amssymb,amsfonts}
\usepackage{graphicx}
\usepackage{textcomp}
\usepackage{xcolor}
\usepackage{multirow}
\usepackage[colorlinks=true, urlcolor=blue, citecolor=blue, linkcolor=darkgray]{hyperref}
\usepackage{bm}

\def\BibTeX{{\rm B\kern-.05em{\sc i\kern-.025em b}\kern-.08em
    T\kern-.1667em\lower.7ex\hbox{E}\kern-.125emX}}

\newcommand{\SensorOneShotStd}{73.5}

\newcommand{\SensorOneShotGcpn}{93.1}
\newcommand{\SensorOneShotGcpnCI}{0.3}
\newcommand{\SensorOneShotGain}{19.6}

\newcommand{\SensorEqualBudgetProto}{93.1}
\newcommand{\SensorEqualBudgetProtoCI}{0.4}
\newcommand{\SensorOneShotMeanAgg}{93.0}

\newcommand{\SensorOneShotMedoid}{91.0}

\newcommand{\SensorFiveShotMeanAgg}{93.9}

\newcommand{\SensorFiveShotMedoid}{93.6}

\newcommand{\SensorNormalStd}{73.1}
\newcommand{\SensorNormalGcpn}{85.9}
\newcommand{\SensorBiasStd}{80.1}
\newcommand{\SensorBiasGcpn}{87.3}

\newcommand{\MnistOneShotStd}{91.6}
\newcommand{\MnistOneShotGcpn}{93.9}
\newcommand{\MnistOneShotGain}{2.3}

\begin{document}

\title{Robust Prototypical Networks for Few-Shot Sensor Fault Diagnosis}

\author{
    \IEEEauthorblockN{%
        Mohammed Ayalew Belay\IEEEauthorrefmark{1}\IEEEauthorrefmark{3},
        Amirshayan Haghipour\IEEEauthorrefmark{2}\IEEEauthorrefmark{3},
        Pierluigi Salvo Rossi\IEEEauthorrefmark{2}\IEEEauthorrefmark{4}%
    }
    \IEEEauthorblockA{%
        \IEEEauthorrefmark{1}Simula UiB, Bergen, Norway \quad
        \IEEEauthorrefmark{2}Dept.\ of Electronic Systems, NTNU, Trondheim, Norway}
    \IEEEauthorblockA{%
        \IEEEauthorrefmark{4}Dept.\ of Gas Technology, SINTEF Energy Research, Trondheim, Norway}
    \IEEEauthorblockA{%
        Emails: ayalew@simula.no, amirshayan.haghipour@ntnu.no, salvorossi@ieee.org}
    \thanks{\IEEEauthorrefmark{3}These authors contributed equally to this work.}
}

\maketitle

\begin{abstract}
Industrial fault diagnosis often operates with only a handful of labeled fault examples,
making few-shot learning attractive for sensor monitoring.
Standard prototypical networks are simple and effective; however, their class prototypes may
become unstable in the very-low-shot regime because each decision relies on a small support set.
We propose \emph{Multi-Episode Prototypical Networks} (MEPN), which aggregate prototypes from
multiple disjoint support episodes and use their mean as the final class representative,
reducing prototype variance without changing the encoder architecture.
We evaluate MEPN on the DeFACTO sensor dataset using five-way fault classification with
synthetic bias, drift, spike, and noise faults injected into real industrial measurements.
Over 100 independent runs, MEPN reaches \textbf{\SensorOneShotGcpn\%} in the per-episode
one-shot setting ($K\!=\!1$ shot, aggregated over $N_{\text{agg}}\!=\!10$ support episodes),
substantially above single-episode
baselines.
Under an equal 10-sample support budget, MEPN and ProtoNet at $K\!=\!10$ are statistically
indistinguishable, confirming prototype accumulation as the mechanism rather than superior
fixed-budget learning.
\end{abstract}


\section{Introduction}
\label{sec:introduction}

Reliable fault detection in industrial sensor networks is difficult when labeled fault examples
are scarce and arrive gradually over time.
Sensor fault diagnosis has traditionally relied on model-based residual analysis, statistical
process monitoring, and PCA; deep learning improves performance through convolutional,
recurrent, and autoencoder architectures~\cite{haghipour2025sensor}, but these assume large
labeled fault libraries that are rarely available in practice.
Few-shot learning provides a natural alternative~\cite{Wang2020GeneralizingLearning,
Song2023AOpportunities}.
Metric-based methods such as prototypical networks~\cite{Snell2017PrototypicalLearning},
Matching Networks~\cite{Vinyals2016MatchingLearning}, and Relation
Networks~\cite{Sung2018LearningLearning} classify new classes without retraining by reasoning
in an embedding space.
Memory-augmented and optimization-based meta-learners pursue fast adaptation through learned
initializations~\cite{Santoro2016Meta-learningNetworks,Ravi2017OptimizationLearning,
Finn2017Model-agnosticNetworks}.
More expressive backbones further add variational autoencoders, adversarial generators,
transformers, and graph networks~\cite{Schonfeld2019GeneralizedAutoencoders,
ZHANG2018MetaGAN:Learning,Han2022Few-ShotCross-Transformer,Garcia2018Few-ShotNetworks};
yet stronger feature extractors do not remove the statistical instability of estimating a
class representative from one or two examples.
Closer to our approach, Kalman Prototypical Networks~\cite{belay2026kalman} stabilize the
prototype itself by modeling it as a latent state updated via Kalman filtering, improving
few-shot fault detection in combined-cycle gas turbines; MEPN pursues a simpler,
filter-free alternative, stabilizing the prototype purely through multi-episode resampling at
inference without any additional dynamical model.

The key limitation is \emph{prototype variance}: if the support set is noisy or
unrepresentative, the class center shifts and all query decisions inherit that error.
This is especially acute for subtle sensor faults with weak class boundaries.
We address it with \emph{Multi-Episode Prototypical Networks} (MEPN), which collect prototypes
across multiple disjoint support episodes and average them into a more stable representative.
Our main contributions are:
(i)~MEPN substantially improves per-episode one-shot accuracy on the DeFACTO sensor benchmark;
(ii)~under the same total support budget it matches single-episode ProtoNet at $K\!=\!10$;
(iii)~an ablation over the aggregation count $N_{\text{agg}}$ and a mean-vs-medoid comparison confirm variance
reduction as the operative mechanism.

\section{Proposed Method}
\label{sec:methodology}

MEPN retains the encoder and distance-based classifier of the standard prototypical
network~\cite{Snell2017PrototypicalLearning} and modifies only the class estimator.
Consider an $n_{\text{way}}$ episode with class set $\mathcal{Y}'$ and embedding
network $f_\theta:\mathcal{X}\!\to\!\mathbb{R}^{D}$.
For class $c$ and support episode $j$ with $K$ labeled examples $\mathcal{S}_c^{(j)}$,
the per-episode prototype is the support centroid in embedding space,
\begin{equation}
\label{eq:proto}
p_c^{(j)}=\frac{1}{K}\sum_{x\in\mathcal{S}_c^{(j)}} f_\theta(x).
\end{equation}
Standard ProtoNet keeps a single prototype per class.
MEPN instead draws $N_{\text{agg}}$ episodes with \emph{disjoint} support sets and
averages their per-class prototypes,
\begin{equation}
\label{eq:agg}
\hat{p}_c=\frac{1}{N_{\text{agg}}}\sum_{j=1}^{N_{\text{agg}}} p_c^{(j)},
\end{equation}
then classifies queries with the unchanged nearest-prototype rule,
\begin{equation}
\label{eq:rule}
\hat{y}(x')=\arg\min_{c}\big\|f_\theta(x')-\hat{p}_c\big\|_2^2.
\end{equation}
At $N_{\text{agg}}=1$ MEPN coincides with ProtoNet.

\subsection{Variance analysis}
\label{subsec:theory}

Modeling class-$c$ support embeddings as i.i.d.\ (independent and identically distributed) draws from a distribution with mean
$\mu_c$ and covariance $\Sigma_c$, the aggregated prototype~\eqref{eq:agg} is unbiased with
\begin{equation}
\label{eq:cov}
\mathrm{Cov}[\hat{p}_c]=\frac{\Sigma_c}{K\,N_{\text{agg}}}.
\end{equation}
MEPN therefore achieves the same prototype covariance as a ProtoNet using
$KN_{\text{agg}}$ shots in a single episode, without requiring all labeled samples to be
available simultaneously.
Sensor segments are extracted from overlapping windows of correlated time series, so the
i.i.d.\ assumption is only approximate.
Modeling the $M\!=\!KN_{\text{agg}}$ pooled support embeddings as equicorrelated with
variance $\sigma_c^2$ and mean pairwise correlation $\bar{\rho}$,
\begin{equation}
\label{eq:floor}
\mathrm{Var}[\hat{p}_c]=\frac{\sigma_c^2}{M}\bigl[1+(M-1)\bar{\rho}\bigr]
\;\xrightarrow[\;M\to\infty\;]{}\;\sigma_c^2\,\bar{\rho}.
\end{equation}
Aggregation still reduces variance monotonically, but toward a floor $\sigma_c^2\bar{\rho}$
rather than zero, predicting the diminishing returns observed beyond $N_{\text{agg}}=10$
(Sec.~\ref{subsec:ablation_nagg}).
For two closely separated classes $a,b$ the relative misorientation of the estimated decision
boundary normal is
$\mathrm{tr}(\Sigma_a+\Sigma_b)\,/\,[K N_{\text{agg}}\,\|\mu_a-\mu_b\|_2^2]$;
small inter-class separation makes this ratio large at low $KN_{\text{agg}}$, precisely where
increasing $N_{\text{agg}}$ removes the most boundary noise.

\subsection{Training and inference}
\label{subsec:training}

During meta-training each update samples one disjoint support/query episode pair with
$N_{\text{agg}}=1$; aggregation is applied only at inference and changes neither the training
objective nor the network parameters.
At inference MEPN collects $N_{\text{agg}}$ support-only episodes, computes one aggregated
prototype per class via~\eqref{eq:agg}, and evaluates on disjoint query episodes.
Enforcing support/query disjointness is essential in the sensor setting, where overlap between
nearby, strongly correlated windows would otherwise inflate accuracy.
Concretely, for each evaluation episode all $N_{\text{agg}}K+Q$ segment indices needed for its
$N_{\text{agg}}$ support rounds and query set are drawn once, without replacement, from the
class's segment pool, so no segment index repeats across support rounds or the query set within
that episode.
This guarantees disjointness in segment \emph{identity} but not in temporal adjacency, since two
distinct, disjointly-indexed windows can still start only a few samples apart and remain
strongly correlated -- precisely the residual correlation $\bar{\rho}$ modeled
in~\eqref{eq:floor}.
As a robustness alternative, the aggregated representative can be set to the medoid of the
collected per-episode prototypes; on the sensor benchmark the lower-variance mean is preferable
and is used as the default throughout.

\section{Experimental Setup}
\label{sec:experiments}

We evaluate on \emph{DeFACTO} (Demonstration of Flow Assurance for CO$_2$ Transport
Operations)~\cite{haghipour2025sensor}, a SINTEF Energy Research facility circulating CO$_2$
through a $90$\,m deep U-tube loop instrumented with over $100$ sensors; we use real industrial
temperature measurements from this facility ($68$ channels, $24{,}988$ time steps at
$2/3$~Hz).
An 80/20 contiguous train/test split avoids temporal leakage; normalization statistics are
computed on the training split only.
We extract 1D segments of length $L=128$ by sampling random start positions independently
within each partition (segments may overlap, but no segment crosses the split boundary).

Each segment belongs to one of five classes.
Faults are injected onto clean standardized segments $\mathbf{s}\in\mathbb{R}^L$ as follows:
\begin{equation}
\label{eq:faults}
\mathbf{x}=
\begin{cases}
\mathbf{s}, & \text{Normal},\\[2pt]
\mathbf{s}+b\,\mathbf{1}, & \text{Bias},\\[2pt]
\mathbf{s}+\mathbf{d},\;\; d_i=\tfrac{i-1}{L-1}d_{\max}, & \text{Drift},\\[2pt]
\mathbf{s}+a\!\!\sum_{i\in\mathcal{M}}\!\mathbf{e}_i,\;\; |\mathcal{M}|=m, & \text{Spike},\\[2pt]
\mathbf{s}+\boldsymbol{\varepsilon},\;\boldsymbol{\varepsilon}\!\sim\!\mathcal{N}(\mathbf{0},\sigma^2\mathbf{I}), & \text{Noise},
\end{cases}
\end{equation}
with parameters $b\!=\!1.5$, $d_{\max}\!=\!1.5$, $m\!=\!2$, $a\!=\!0.6$, $\sigma\!=\!0.06$, chosen
to be conservative relative to the fault severities typically considered detectable in
industrial sensor-validation practice~\cite{haghipour2025sensor}.
The deliberately small magnitudes produce close class centres in embedding space; by~\eqref{eq:cov}
this is exactly the regime where prototype variance dominates one-shot error, and where a
weaker, harder-to-catch synthetic fault is the more realistic diagnostic target.
We sample $10{,}000$ training and $5{,}000$ test segments.

All methods share the same 1D-CNN encoder (four convolutional blocks, 64-dim embedding) and
episodic training protocol~\cite{Snell2017PrototypicalLearning,Chen2019AClassification}.
Training uses Adam with learning rate $10^{-3}$.
Each model is meta-trained for $200$ episodic iterations using $5$-way episodes with $15$ query
examples per class; no learning-rate schedule, weight decay, or early stopping is used, and no
method-specific hyperparameter search was performed, so all baselines share this configuration.
Training and inference use PyTorch on a single GPU when available (CPU otherwise).
MEPN uses $N_{\text{agg}}=10$ at inference.
Each of the 100 independent runs uses a fresh random seed that reseeds segment sampling (new
$10{,}000$/$5{,}000$ train/test draws), model initialization, and training; results are averaged
over these 100 runs, with tables reporting mean accuracy and 95\% confidence intervals computed
as $1.96\,\hat{\sigma}/\sqrt{100}$.

\section{Results and Discussion}
\label{sec:results}

\begin{figure}[t]
  \centering
  \includegraphics[width=0.92\linewidth]{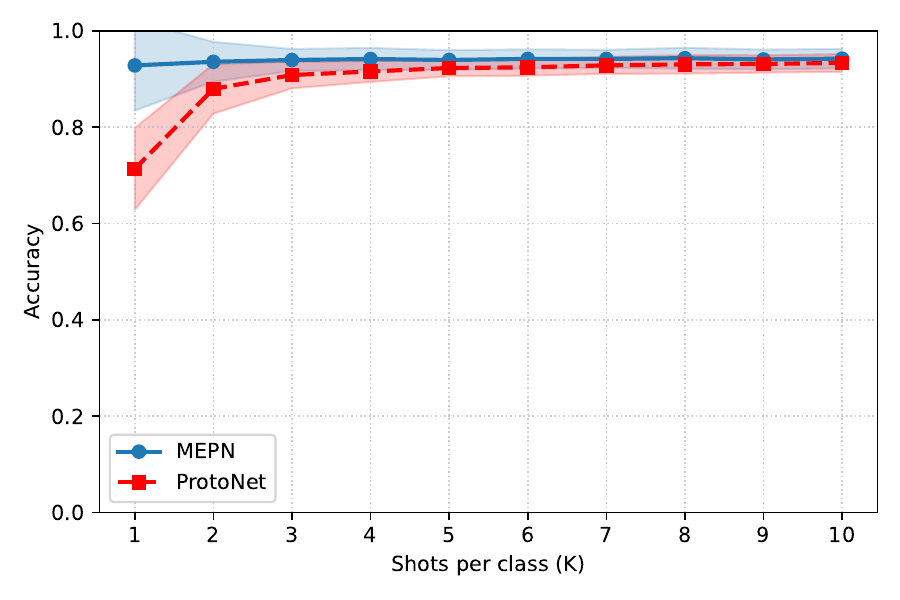}
  \caption{Mean five-way classification accuracy (vertical axis in $[0,1]$, i.e., fraction
           correct) vs.\ support examples per class ($K\!=\!1\!\dots\!10$) on the DeFACTO
           dataset, averaged over 100 runs. Shaded regions: mean~$\pm$~95\%~CI.}
  \label{fig:acc_vs_shots}
\end{figure}

\textbf{Accuracy vs.\ shots.}
\autoref{fig:acc_vs_shots} reports mean five-way accuracy as $K$ varies from 1 to 10.
MEPN is consistently stronger in the low-shot regime: at $K\!=\!1$ it reaches
\SensorOneShotGcpn\%, versus the single-episode ProtoNet baseline of \SensorOneShotStd\%
(a gain of \SensorOneShotGain{} percentage points).
Performance saturates around five shots, where both methods become more stable and the gap
narrows, consistent with~\eqref{eq:cov}.

\textbf{Comparison with competing methods.}
\autoref{tab:comparison} compares MEPN with ProtoNet, Matching Networks, and Relation Networks
under the same encoder and protocol.
MEPN is best at both $K\!=\!1$ and $K\!=\!5$, with the largest margin at $K\!=\!1$ where
prototype instability is most severe.
Matching Networks employ instance-level attention, which reduces the sensitivity associated
with one-shot learning and improves performance over the single-episode ProtoNet; however, they
still perform below MEPN because they do not explicitly reduce support-side prototype variance.
Relation Networks fare worst: a learned relation module alone is insufficient when the support
signal is weak and faults are only subtly separated.

\begin{table}[t]
\centering
\caption{Five-way $K$-shot fault classification accuracy (\%) on the DeFACTO sensor dataset. Mean\,$\pm$\,95\% CI over 100 independent runs. The $K=1$ and $K=5$ columns match methods by per-episode shot count; MEPN additionally aggregates $N_{\text{agg}}=10$ disjoint support episodes per class, so these columns are not fixed-budget comparisons. The bottom row is the direct equal-budget reference. Best result per column in \textbf{bold}.}
\label{tab:comparison}
\renewcommand{\arraystretch}{1.2}
\begin{tabular}{lcc}
\hline
\textbf{Method} & $K=1$ & $K=5$ \\
\hline
ProtoNet & 73.5 $\pm$ 1.6 & 92.2 $\pm$ 0.4 \\
Matching Networks & 77.8 $\pm$ 1.0 & 81.7 $\pm$ 0.9 \\
Relation Networks & 48.9 $\pm$ 2.5 & 58.5 $\pm$ 2.4 \\
MEPN (proposed) & \textbf{93.1} $\pm$ 0.3 & \textbf{93.8} $\pm$ 0.3 \\
\hline
ProtoNet (K=10, equal budget) & \multicolumn{2}{c}{93.1 $\pm$ 0.4} \\
\hline
\end{tabular}
\end{table}

The bottom row of \autoref{tab:comparison} addresses support-budget fairness.
MEPN at $K\!=\!1$, $N_{\text{agg}}\!=\!10$ consumes the same total support samples as
ProtoNet at $K\!=\!10$.
Accuracies are \SensorOneShotGcpn\%\,$\pm$\,\SensorOneShotGcpnCI\% and
\SensorEqualBudgetProto\%\,$\pm$\,\SensorEqualBudgetProtoCI\% respectively, with fully
overlapping 95\% CIs.
MEPN therefore does not create information but makes a fixed budget usable incrementally,
without waiting for all labels to arrive at once.

\subsection{Confusion matrix analysis}
\label{subsec:confusion}

\begin{figure}[t]
  \centering
  \includegraphics[width=0.49\linewidth]{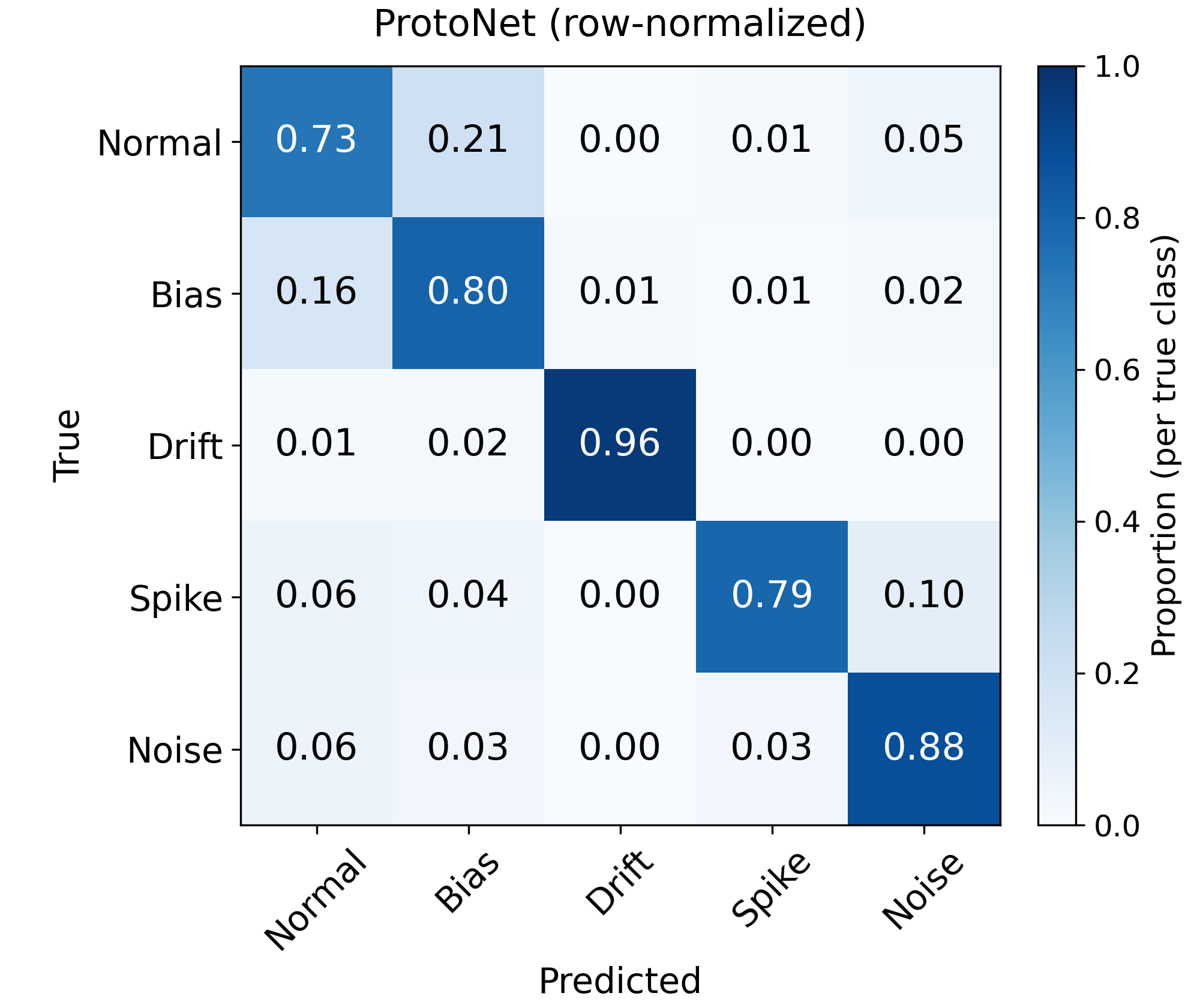}%
  \hfill
  \includegraphics[width=0.49\linewidth]{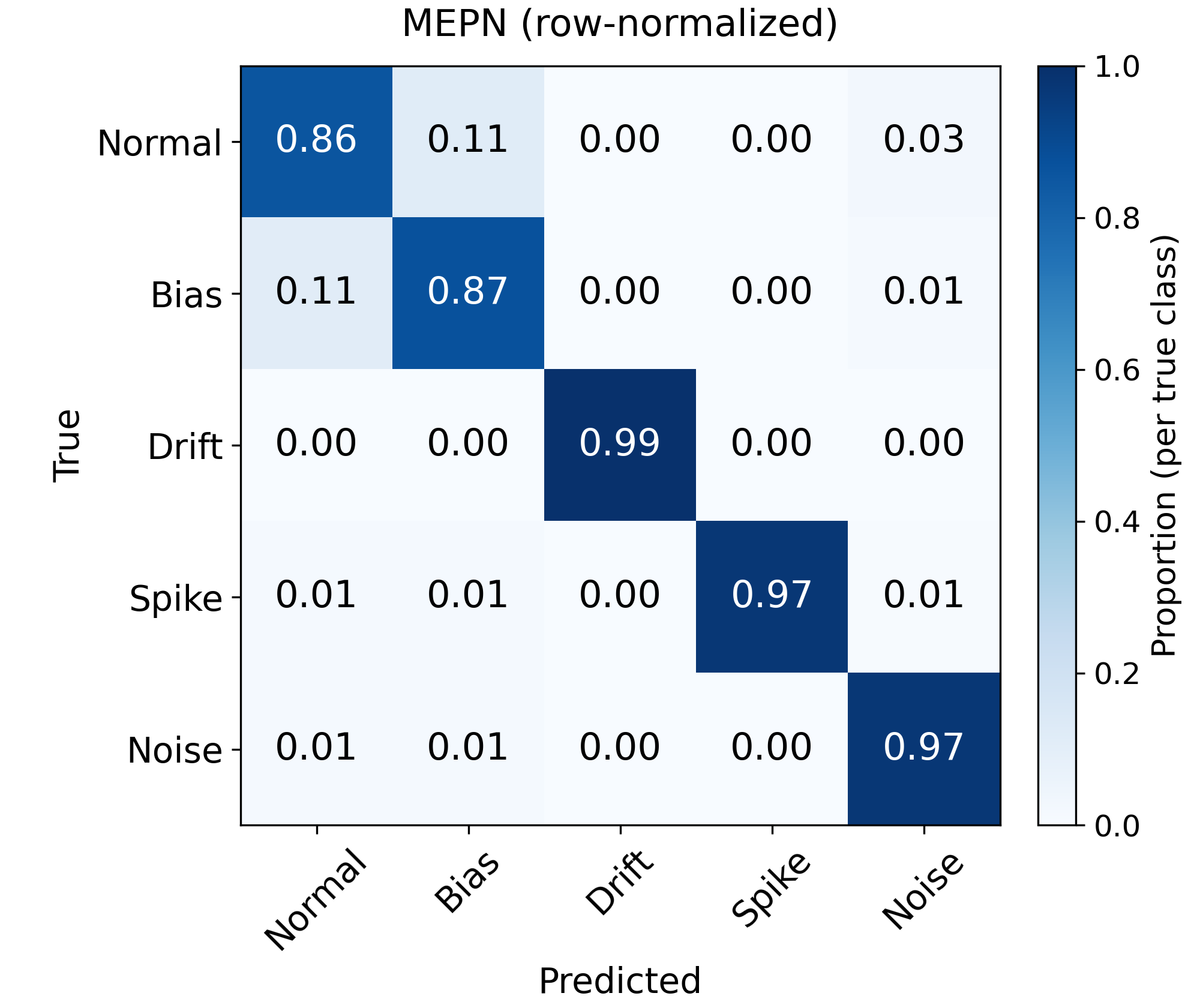}
  \caption{Mean row-normalized confusion matrices at $K\!=\!1$, averaged over 100 runs.
           Left: ProtoNet. Right: MEPN (proposed).
           Prototype aggregation reduces Normal/Bias confusion and resolves Spike/Noise overlap.}
  \label{fig:confusion}
\end{figure}

\autoref{fig:confusion} shows the mean row-normalized confusion matrices at $K\!=\!1$.
The dominant failure mode of the standard baseline is confusion between Normal and Bias,
while Spike and Noise remain partially entangled because both manifest as short local
disturbances.
MEPN improves the true-positive rate for Normal from \SensorNormalStd\% to
\SensorNormalGcpn\% and for Bias from \SensorBiasStd\% to \SensorBiasGcpn\%,
directly reducing the most operationally costly error.
Spike and Noise also improve substantially, while Drift remains the easiest class for both
methods because its full-window trend is structurally distinct from all other fault types.

\subsection{Ablation: Effect of \texorpdfstring{$N_{\text{agg}}$}{Nagg}}
\label{subsec:ablation_nagg}

\autoref{tab:ablation_nagg} shows MEPN accuracy vs.\ aggregation count.
Moving from $N_{\text{agg}}\!=\!1$ to $2$ already raises $K\!=\!1$ from $73.5\%$ to $89.1\%$;
gains continue through $N_{\text{agg}}\!=\!10$ then saturate, consistent with the
variance-floor of~\eqref{eq:floor}.
The confidence interval shrinks monotonically, so MEPN is both more accurate and more
reliable as $N_{\text{agg}}$ grows.

\begin{table}[t]
\centering
\caption{Ablation: MEPN accuracy (\%) vs.\ $N_{\text{agg}}$ on DeFACTO, from an independent
         100-run batch (different random-seed sequence than \autoref{tab:comparison}); the
         $N_{\text{agg}}\!=\!1$ row and the ProtoNet reference below therefore differ slightly
         from \autoref{tab:comparison} (e.g., $71.3\%$ vs.\ $73.5\%$ at $K\!=\!1$) but agree
         within the reported 95\% CIs.
         Mean $\pm$ 95\% CI over 100 runs.
         ProtoNet reference (unchanged by $N_{\text{agg}}$):
         $K\!=\!1$: $71.3\pm1.6$;\; $K\!=\!5$: $91.8\pm0.5$.}
\label{tab:ablation_nagg}
\renewcommand{\arraystretch}{1.2}
\begin{tabular}{ccc}
\hline
$N_{\text{agg}}$ & $K=1$ & $K=5$ \\
\hline
1  & 73.5 $\pm$ 1.7 & 91.9 $\pm$ 0.4 \\
2  & 89.1 $\pm$ 0.7 & 93.1 $\pm$ 0.4 \\
5  & 92.3 $\pm$ 0.3 & 93.9 $\pm$ 0.4 \\
10 & 93.0 $\pm$ 0.3 & 93.9 $\pm$ 0.3 \\
20 & 93.8 $\pm$ 0.3 & 94.1 $\pm$ 0.3 \\
\hline
\end{tabular}
\end{table}

\subsection{Mean vs.\ medoid aggregation}
\label{subsec:mean_vs_medoid}

\autoref{tab:mean_vs_medoid} compares mean and medoid aggregation at $N_{\text{agg}}\!=\!10$.
Mean aggregation outperforms the medoid at both shot counts, reaching \SensorOneShotMeanAgg\%
vs.\ \SensorOneShotMedoid\% at $K\!=\!1$ and \SensorFiveShotMeanAgg\%
vs.\ \SensorFiveShotMedoid\% at $K\!=\!5$.
The per-episode prototypes cluster tightly enough that the outlier-resistance of the medoid
is not needed; the lower-variance mean of~\eqref{eq:cov} is uniformly preferable,
confirming that \emph{averaging} rather than central-candidate selection drives the improvement.

\begin{table}[t]
\centering
\caption{Mean aggregation vs.\ medoid selection on the DeFACTO sensor dataset. Mean\,$\pm$\,95\% CI over 100 runs ($N_{\text{agg}}=10$; same rerun batch as \autoref{tab:ablation_nagg}, hence the ProtoNet baseline here also differs slightly from \autoref{tab:comparison}). Both MEPN variants use the same multi-episode support collection; the only difference is the final aggregation step.}
\label{tab:mean_vs_medoid}
\renewcommand{\arraystretch}{1.2}
\begin{tabular}{lcc}
\hline
\textbf{Method} & $K=1$ & $K=5$ \\
\hline
MEPN (medoid variant) & 91.0 $\pm$ 0.5 & 93.6 $\pm$ 0.3 \\
MEPN (mean, proposed) & \textbf{93.0} $\pm$ 0.3 & \textbf{93.9} $\pm$ 0.3 \\
ProtoNet (baseline) & 71.3 $\pm$ 1.6 & 91.8 $\pm$ 0.4 \\
\hline
\end{tabular}
\end{table}

We also validated MEPN on MNIST as an image-domain sanity check, where it improves one-shot
accuracy from \MnistOneShotStd\% to \MnistOneShotGcpn\% (a gain of \MnistOneShotGain{}~pp),
with the gap narrowing as $K$ increases, confirming that the benefit addresses a general
weakness of single-episode prototype estimation rather than a peculiarity of DeFACTO.

\section{Conclusions}
\label{sec:conclusions}

MEPN improves few-shot sensor fault classification by replacing single-episode prototypes with
multi-episode aggregated prototypes.
On DeFACTO it raises one-shot accuracy to \SensorOneShotGcpn\% and matches a ProtoNet that
spends the same total support budget in a single batch; confusion-matrix analysis confirms
the gain is operationally meaningful, most prominently by reducing Normal/Bias ambiguity.
The improvement saturates near $N_{\text{agg}}\!=\!10$, with mean aggregation uniformly
beating the medoid, in line with~\eqref{eq:cov}--\eqref{eq:floor}.
These results indicate that the design of the class \emph{estimator} is as important as that of
the encoder, particularly in safety-critical industrial monitoring applications where
prediction stability is essential.

\bibliographystyle{IEEEtran}
\bibliography{ref}

\end{document}